\PassOptionsToPackage{dvipsnames,table}{xcolor}
\documentclass[sigconf=true, nonacm=true, review=false, anonymous = false]{acmart}
\AtBeginDocument{%
  \providecommand\BibTeX{{%
    \normalfont B\kern-0.5em{\scshape i\kern-0.25em b}\kern-0.8em\TeX}}}

\begin{document}

\title{Assessing the Generalizability of a Performance Predictive Model}

\author{Ana Nikolikj}
\orcid{0000-0002-6983-9627}
\affiliation{
  \institution{Jo\v{z}ef Stefan Institute \&}
  \institution{Jo\v{z}ef Stefan International Postgraduate School}
  \streetaddress{Jamova cesta 39}
  \city{Ljubljana}
   \country{Slovenia}
  \postcode{1000 }
}

\author{Gjorgjina Cenikj}
\orcid{0000-0002-2723-0821}
\affiliation{
   \institution{Jo\v{z}ef Stefan Institute \&}
  \institution{Jo\v{z}ef Stefan International Postgraduate School}
  \streetaddress{Jamova cesta 39}
  \city{Ljubljana}
   \country{Slovenia}
  \postcode{1000 }
}

\author{Gordana Ispirova}
\orcid{0000-0001-8558-6909}
\affiliation{
  \institution{Jo\v{z}ef Stefan Institute}
  \streetaddress{Jamova cesta 39}
  \city{Ljubljana}
   \country{Slovenia}
  \postcode{1000 }
}

\author{Diederick Vermetten}
\orcid{0000-0003-3040-7162}
\affiliation{
  \institution{LIACS, Leiden University}
  \country{The Netherlands}
}

\author{Ryan Dieter Lang}
\orcid{0000-0001-5051-5414}
\affiliation{
  \institution{Stellenbosch University}
  \country{South Africa}
}

\author{Andries Petrus Engelbrecht}
\orcid{0000-0002-0242-3539}
\affiliation{
  \institution{Stellenbosch University}
  \country{South Africa}
}
\additionalaffiliation{
    \institution{Gulf University for Science and Technology}
}

\author{Carola Doerr}
\orcid{0000-0002-5983-7169}
\affiliation{%
  \institution{Sorbonne Université, CNRS, LIP}
  \city{Paris}
   \country{France}
  \postcode{1000}
}

\author{Peter Koro\v{s}ec}
\orcid{0000-0003-4492-4603}
\affiliation{
  \institution{Jo\v{z}ef Stefan Institute}
  \country{Slovenia}
}

\author{Tome Eftimov}
\orcid{0000-0001-7330-1902}
\affiliation{
  \institution{Jo\v{z}ef Stefan Institute}
  \country{Slovenia}
}

\renewcommand{\shortauthors}{Nikolikj, et al.}

\begin{abstract}
A key component of automated algorithm selection and configuration, which in most cases are performed using supervised machine learning (ML) methods is a good-performing predictive model. The predictive model uses the feature representation of a set of problem instances as input data and predicts the algorithm performance achieved on them. Common machine learning models struggle to make predictions for instances with feature representations not covered by the training data, resulting in poor generalization to unseen problems. In this study, we propose a workflow to estimate the generalizability of a predictive model for algorithm performance, trained on one benchmark suite to another. The workflow has been tested by training predictive models across benchmark suites and the results show that generalizability patterns in the landscape feature space are reflected in the performance space. 
\end{abstract}



\maketitle

\section{Introduction}
Automated algorithm configuration~\cite{prager2020per,BelkhirDSS17} and selection~\cite{jankovic2020landscape,kerschke2019automated} are among the most researched topics in evolutionary computation. 
These systems often use predictive machine learning (ML) models which take the feature representation of problem instances as input and predict the performance of an algorithm instance on a problem instance. 
However, one of the main drawbacks presented in these learning tasks is the low generalizability of the predictive models. 
The models fail to provide accurate predictions for problem instances whose feature representation is underrepresented or not presented in the training data.

Recent studies~\cite{vskvorc2022transfer,kostovska2022per} show that poor predictive results have been obtained when an ML model for performance prediction is trained on the problem instances from one benchmark suite and then evaluated on problem instances from another benchmark suite. 
\v{S}kvorc et al.~\cite{vskvorc2022transfer} present results when a random forest (RF) model trained on the BBOB (i.e., COCO)~\cite{hansen2021coco} benchmark suite provides poor results when tested on artificially generated problem instances~\cite{tian2020recommender} and vice-versa. 
Kostovska et al.~\cite{kostovska2022per} show that an automated algorithm selector which is based on performance prediction models trained on the BBOB benchmark suite, cannot generalize to problem instances from the Nevergrad's YABBOB~\cite{bennet2021nevergrad} benchmark suite.

\textbf{Our contribution:} We propose a workflow to estimate the generalizability of a predictive model from one benchmark suite to another. 
Problem instances are grouped into clusters based on their features and further use the distribution of each benchmark suite represented as the number of problem instances across the clusters as a meta-representation for each benchmark suite. 
Similarity between benchmark suites is calculated using this meta-representation. 
This similarity can indicate if a predictive model can be generalized across benchmark suites. 
We evaluated the workflow by training predictive models and found that generalizability patterns in the feature space were also present in the performance space.

\textbf{Data and code availability:} The data and the code involved in this study are available at ~\cite{gitGenerazibility}.

\section{Assessing generalizability workflow}
\label{sec:gen_workflow}
Let us assume that we have $m$ benchmark suites. Each benchmark suite can consist of a different number of problem instances. One out of $m$ benchmark suites is selected to train the supervised ML predictive model ($\mathcal{M}$) and the remaining $m-1$ benchmark suites are used to test the model. To assess the generalizability of the model $\mathcal{M}$ to the different benchmark suites used for testing, we propose the following workflow:

\noindent\textbf{Defining a unified meta-representation on a problem instance level} -- represent the problem instances from all benchmark suites using the same $n$ meta-features that describe the landscape properties of the problem instances. With this, all problem instances (i.e., the ones selected for training and the remaining ones used for testing) are projected into the same $n$-dimensional vector space.
\noindent\textbf{Defining a coverage matrix} -- based on their meta-representation cluster the problem instances from all benchmark suites into $k$ clusters. Next, for each benchmark suite calculate the percentage of problem instances that belong to each cluster. With this, we define $k$-vector meta-representation on a benchmark suite level that represents the distribution of the benchmark suite across different clusters (i.e different regions in the problem space).

\noindent\textbf{Define the similarity between two benchmark suites} -- the similarity between a pair of benchmark suites is calculated using their coverage matrix-based meta-representation. The approach uses cosine similarity as a similarity measure~\cite{singhal2001modern}.

High benchmark suite similarity suggests accurate predictions by a model trained on one suite for the other suite. The low similarity suggests poor generalization and coverage of different problem landscape space regions.

\vspace{-1.5mm}

\section{Experimental design}
\label{sec:experiments}
The evaluation of the workflow has been performed in two experiments. More details about them are provided below.

\noindent\textbf{Benchmark suites}: 
In the \emph{first experiment}, we involve the benchmark suite data available from a previous study~\cite{lang2021exploratory}, where the BBOB, CEC 2013, CEC 2014, CEC 2015, and CEC 2017 benchmark suites are used. 
The CEC benchmark suites change annually, with some overlap in problem instances across different suites, but the definition of the same problem instance differs each year, which may result in varying properties of the benchmarks even with the same problem instance definition. 
In the experiments, the problem dimension is set to $D = 10$. 
In the \emph{second experiment}, we select benchmark problem instances that are affine recombinations of pairs of BBOB problem instances, where 9,936 new problem instances are generated for several dimensions~\cite{dietrich2022increasing}. 
Next, we use the SELECTOR approach~\cite{selector} to select diverse benchmark problem instances in $D=5$ based on their 14 landscape features. The benchmark problem instances have been transformed into a graph based on the cosine similarity using their landscape features. Next, the Maximal Independent Set method has been run five times independently to select five benchmark suites (BS1, BS2, BS3, BS4, BS5) that contain around 110 problem instances with minimal overlap. SELECTOR guarantees that the distribution of the problem instances in the five independent selections is similar.

\noindent\textbf{Performance data:} In the \emph{first experiment}, performance data for the Covariance Matrix Adaption Evolutionary Strategy (CMA-ES)~\cite{hansen2001self_adaptation_es} has been used.  The algorithm stops after either reaching 100,000 function evaluations or finding a solution within $10^{-8}$ of the global optimum. As a target in the regression models, we used the obtained solutions' precision (i.e., the error to the global optimum). In the \emph{second experiment}, we use the performance data of the Diagonal CMA-ES~\cite{hansen2001self_adaptation_es}. Here, we also retrieve the precision after a budget of 10,000 function evaluations as a target variable for the regression models. For both experiments, we use their default hyper-parameters implementation from the Nevergrad library~\cite{bennet2021nevergrad}.

\noindent\textbf{Exploratory landscape analysis:} For the first experiment, to describe the landscape properties of each problem instance, 64 publicly available ELA features  from a previous study~\cite{lang2021exploratory} are used. In the second experiment, 14 ELA features are used, also available from a previous study~\cite{dietrich2022increasing}.

\noindent\textbf{Clustering:} The K-Means algorithm clusters problem instances from benchmark suites in both experiments. The Silhouette score is used to estimate cluster number in the \emph{first experiment} and the elbow method with the distortion metric is used in the \emph{second experiment}. We tested different measures to estimate the number of clusters, just to check the sensitivity of the approach using different measures. ELA features are normalized before clustering, and the \emph{scikit-learn} package in Python is used for its implementation.

\noindent\textbf{Predictive models:} Random Forest (RF) models (from the \emph{scikit-learn package in Python} with default hyper-parameters) are trained on each benchmark suite, evaluated on remaining suites, and reported using median absolute error (MDAE). Results are analyzed to determine if a pattern from the coverage matrix is also present in automated algorithm performance prediction model performance.

\section{Results and discussion}
\label{sec:results}
Here, the results for both experiments are presented in more detail.

\noindent\textbf{First experiment.} The optimal number of clusters has been estimated to 13. Table~\ref{tab:covMat_13_clust} presents the coverage matrix, where each cell in the table shows the percentage of the total number of instances from the benchmark suite presented in the row, that belongs to each cluster presented in the column. Each row then is used as a meta-representation for each benchmark suite.
The results show that the BBOB benchmark suite is the most widely spread in the feature space (i.e., landscape space) as its instances are distributed across nine clusters out of 13, compared to the CEC benchmark suites which condensed to a smaller number of clusters. There are four clusters that consist only of BBOB problem instances. Also, it is visible that the distribution of the CEC 2014 and CEC 2017 problem instances across the clusters is very similar.

\begin{table}[ht]
\vspace{-3mm}
\centering
\caption{Coverage matrix calculated with 13 clusters.}
\vspace{-4mm}
\label{tab:covMat_13_clust}
\resizebox{.45\textwidth}{!}{
\begin{tabular}{lrrrrrrrrrrrrr}
  \hline
 & C1 & C2 & C3 & C4 & C5 & C6 & C7 & C8 & C9 & C10 & C11 & C12 & C13 \\ 
  \hline
BBOB & 0.04 & 0.47 & 0.08 & 0.00 & 0.03 & 0.08 & 0.06 & 0.00 & 0.00 & 0.00 & 0.18 & 0.04 & 0.02 \\ 
  CEC2013 & 0.00 & 0.00 & 0.00 & 0.40 & 0.04 & 0.28 & 0.04 & 0.08 & 0.08 & 0.00 & 0.00 & 0.08 & 0.00 \\ 
  CEC2014 & 0.00 & 0.00 & 0.03 & 0.30 & 0.10 & 0.17 & 0.10 & 0.27 & 0.00 & 0.03 & 0.00 & 0.00 & 0.00 \\ 
  CEC2015 & 0.00 & 0.00 & 0.00 & 0.13 & 0.00 & 0.13 & 0.27 & 0.33 & 0.00 & 0.07 & 0.00 & 0.07 & 0.00 \\ 
CEC2017 & 0.00 & 0.00 & 0.00 & 0.31 & 0.07 & 0.07 & 0.21 & 0.24 & 0.00 & 0.03 & 0.00 & 0.07 & 0.00 \\ 
   \hline
\end{tabular}
}
\vspace{-3mm}
\end{table}

\begin{figure}
    \centering
    \includegraphics[scale=0.35]{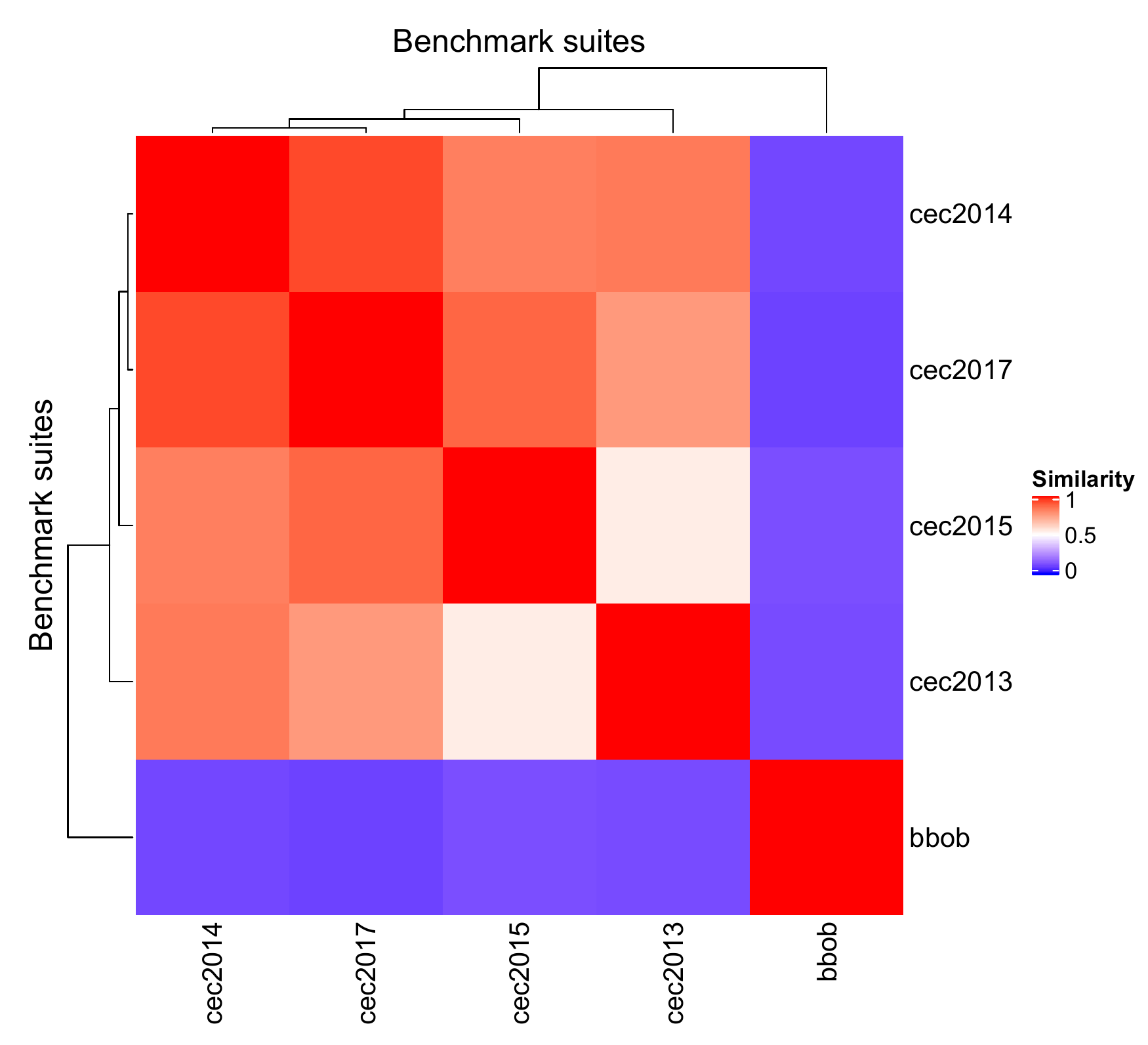}
    \vspace{-5mm} 
    \caption{Heatmap of the cosine similarity between benchmark suites representations generated based on 13 clusters.}
    \vspace{-6mm} 
    \label{fig:heatmap_13_similarity}
\end{figure}

Figure~\ref{fig:heatmap_13_similarity} presents a heatmap of the cosine similarity between the benchmark suite meta-representations, and hierarchical clustering dendrogram of the benchmark suites' meta-representations. Based on the cosine similarity between the benchmark suites, the similarity matrix is reorganized such that the more similar benchmark suites are placed together in the dendrogram. The colors in the plot represent the cosine similarity. The figure  shows that all CEC benchmark suites have high similarity. The pairwise cosine similarities between all pairs of CEC benchmark suites are greater than 0.5. This is as expected since the CEC suites have had little modifications through the years. However, comparing them to the BBOB benchmark suite it seems that there is a big difference in how their problem instances are distributed in the feature space, also visible earlier from the coverage matrix. 
Looking into the clustering result, it follows that CEC 2014 and CEC 2017 are the most similar ones, further both of them are close to CEC2015, and then to CEC 2013, while all of them are placed on another side of the dendrogram compared to the BBOB. This result further points out that we can expect a predictive model trained on CEC 2014 to have the best results when it will be evaluated on CEC 2017 and vice-versa. 
Further, a model trained on CEC 2014 or CEC 2017 is expected to have good prediction results when it will be evaluated on CEC 2013 and CEC 2015. 
Also, models trained on CEC 2013 or CEC 2015 will have to generalize the prediction results across CEC benchmark suites. 
The dissimilarity of CEC benchmark suites with the BBOB benchmark suite indicates that we do not have a guarantee that the model will generalize across them.

\begin{figure}
    \centering
    \includegraphics[trim={0 0 0 50},clip,scale=0.55]{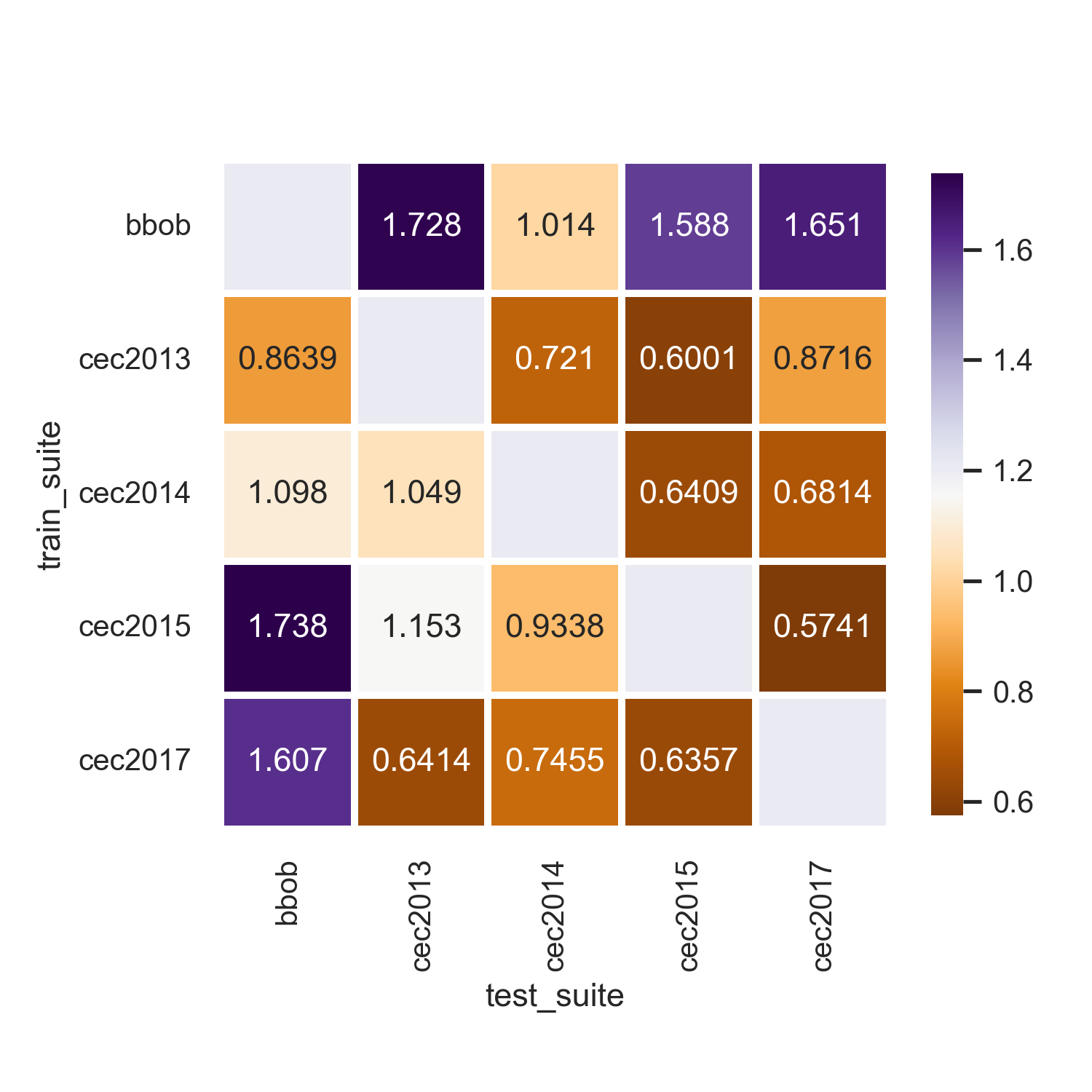}
    \vspace{-9mm} 
    \caption{
        Heatmap showing the MDAEs of an RF model for the performance of CMA. Rows indicate the train benchmark suite and columns indicate the benchmark suite the model was evaluated on.
    }
    \label{fig:CMA_ES}
    \vspace{-5mm}
\end{figure}

To investigate if the similarity patterns in the landscape feature space will also be present in the performance space as model generalizability patterns, we present the evaluation results of an algorithm performance predictive model, when the model is trained on one benchmark suite and evaluated on the remaining ones. The heatmap in Figure~\ref{fig:CMA_ES} presents the RF model errors for the performance prediction of the CMA-ES. Each cell presents the median absolute error of the RF model, trained on the benchmark suite presented in the row, and evaluated on the benchmark suite presented in the column. The results show that a predictive model trained on BBOB produces larger errors across all CEC benchmark suites. A model trained on CEC 2017 provides smaller errors when it is evaluated on CEC 2013, 2014, and 2015, and ends up with a larger error for BBOB. When CEC 2013 is used to train the model, similar errors are obtained across all benchmark suites. A similar effect occurs when CEC 2015 is used for training, good errors are achieved on CEC 2014 and CEC 2017, and the error increases for CEC 2013, ending up with a larger error for BBOB. 

The results indicate that a similar distribution of the benchmark suites over the landscape feature space leads to similar model errors on the suites. This study does not guarantee that the training and testing error will be good but it guarantees that they will be in similar ranges. We are not dealing with the quality of the benchmark suites but only comparing the landscape feature distribution of the benchmark suites. However, it is not possible to establish a complete generalizability mapping function between the landscape feature space and the performance space, since these algorithms are stochastic in nature and they all have different behavior.

\noindent\textbf{Second experiment:} Using the elbow method of the distortion metric curve, the optimal number of clusters is determined to be four. Table~\ref{tab:covMat_4_clust} presents the distribution of the five artificial benchmark suites across the four clusters. From the table, it follows that all benchmark suites have similar distribution across the clusters, thus following the same distribution across the feature space. In addition, for each pair of benchmark suites we analyzed the cosine similarity 
between their meta-representations based on the coverage matrix. The obtained cosine similarities of the meta-representations exceeded 0.98 for all pairs of benchmark suites. This result indicates that a model trained on any one of these benchmark suites should be easy to generalize to the remaining benchmark suites.

\begin{table}[ht]
\vspace{-3mm}
\centering
\caption{Coverage matrix calculated with four clusters.}
\vspace{-4mm}
\label{tab:covMat_4_clust}
\begin{tabular}{rrrrr}
  \hline
 & C1 & C2 & C3 & C4 \\ 
  \hline
BS1 & 0.14 & 0.21 & 0.32 & 0.32 \\ 
  BS2 & 0.16 & 0.26 & 0.33 & 0.25 \\ 
  BS3 & 0.16 & 0.27 & 0.32 & 0.25 \\ 
  BS4 & 0.20 & 0.22 & 0.31 & 0.27 \\ 
  BS5 & 0.15 & 0.26 & 0.34 & 0.25 \\ 
   \hline
\end{tabular}
\vspace{-3mm}
\end{table}

\begin{figure}
    \centering
    \includegraphics[scale=0.2]{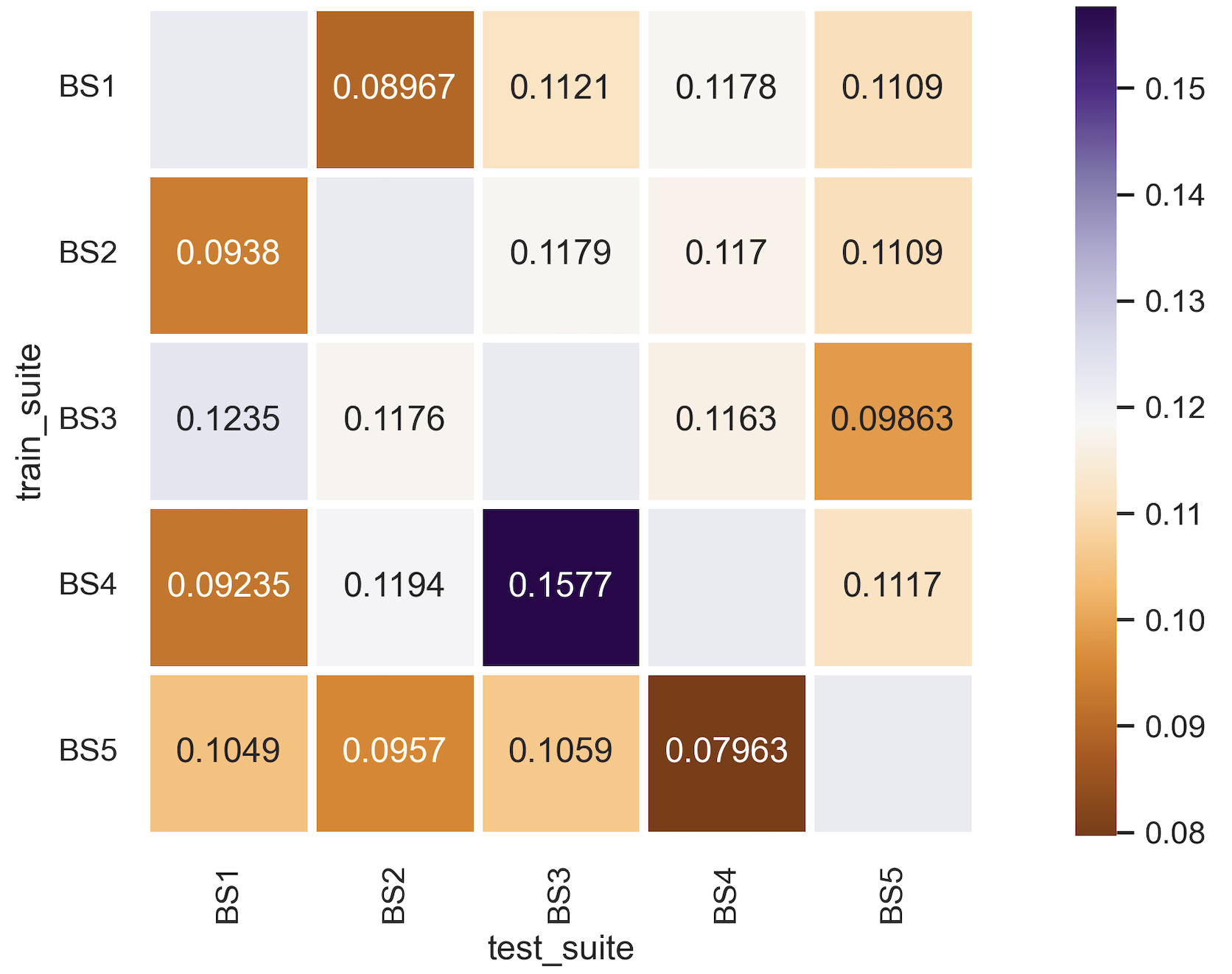}
    \vspace{-4mm}
    \caption{Heatmap showing the MDAEs of an RF model for the performance of diagonal CMA-ES. Rows indicate the train benchmark suite and columns indicate the benchmark suite the model was evaluated on.}
    \label{fig:CMAES_affine}
    \vspace{-5mm}
\end{figure}

The evaluation results (MDAE) of the predictive models trained on each of the five artificial benchmark suites and evaluated on the remaining four for the diagonal CMA-ES are presented in Figure~\ref{fig:CMAES_affine}. The rows indicate the benchmark suite on which the model has been trained and the columns indicate the benchmark suite on which the model has been evaluated.

The results show that the patterns that are visible in the landscape feature space are also reflected in the model performance space. Models trained on all five benchmark suites separately are generalizable or have similar errors on all the remaining benchmarking suites. We also need to point out here that the training MDAE is smaller than all the MDAE obtained on the test benchmark suites, which is expected in a machine learning setup. However, the difference between the training and testing errors is not practically significant (e.g., training MDAE is 0.06 for BS1, 0.04 for BS2, 0.08 for BS3, 0.03 for BS4, and 0.04 for BS5 for diagonal CMA-ES).

To show that different distributions in the feature landscape space lead to worse model performance,we assessed all instances in the fourth cluster. This cluster, referred to as BS6, includes samples from all five artificially generated benchmark suites and was randomly selected to ensure a different feature landscape distribution than the other five suites. The BS6 instances cover only one region of the feature landscape and do not include samples from the three other regions present in the clustering result that are part of the other five suites. Table~\ref{tab:BS6} presents RF errors when the model is trained on BS6 and evaluated for automated algorithm performance prediction on the other five benchmark suites. It is obvious that the error models are worse (compared with the errors presented in Figure~\ref{fig:CMAES_affine}. The results prove that different feature landscape distribution decreases the generalization of a predictive model.

\begin{table}[!htp]
\vspace{-2mm}
\centering
\caption{RF errors when the model is trained on BS6 and evaluated on the other five benchmark suites.}\label{tab:BS6}
\vspace{-4mm}
\scriptsize
\begin{tabular}{lrrrrrr}\toprule
&BS1 &BS2 &BS3 &BS4 &BS5 \\\midrule
DE &0.185922 &0.182945 &0.183876 &0.185423 &0.202225 \\
RSPSO &0.479039 &0.492742 &0.474655 &0.506378 &0.517526 \\
diag CMA-ES &0.240901 &0.240292 &0.237267 &0.25774 &0.240292 \\
\bottomrule
\end{tabular}
\vspace{-3mm}
\end{table}

\section{Conclusion}
\label{sec:conclusion}
Our study proposes a workflow for estimating the generalizability of performance predictive models in automated algorithm selection and configuration. The workflow involves converting problem instances into a common meta-representation and clustering them to create a benchmark suite meta-representation. The similarity between benchmark suites is then calculated to indicate generalizability between models. Two experiments were conducted, one with commonly used benchmark suites and the other with artificially generated suites. Results show that generalizability patterns in the feature landscape space also exist in the performance space, assisting in predicting model performance on new instances. However, the workflow is dependent on the quality of the feature representation and may be affected by different performances of the algorithm with similar feature representations. Future work will test the workflow with different feature representations and families of supervised machine learning methods.

\begin{acks}
The authors acknowledge the support of the Slovenian Research Agency through program grant P2-0098, project grants N2-0239 to TE and J2-4460 to PK, young researcher grant No. PR-12393 to GC,  and a bilateral project between Slovenia and France grant No. BI-FR/23-24-PROTEUS-001 (PR-12040). Our work is also supported by ANR-22-ERCS-0003-01 project VARIATION. The authors also acknowledge the Centre for High Performance Computing (CHPC), South Africa, for providing computational resources to this research project.

\end{acks}


\end{document}